\documentclass{article}

\usepackage[sglblindworkshop, final]{neurips_2026}
\workshoptitle{Agents in the Wild}
\usepackage{amsmath}
\usepackage{textgreek}
\usepackage{upquote}

\usepackage[utf8]{inputenc} 
\usepackage[T1]{fontenc}    
\usepackage{hyperref}       
\usepackage{url}            
\usepackage{booktabs}       
\usepackage{amsfonts}       
\usepackage{nicefrac}       
\usepackage{microtype}      
\usepackage{xcolor}         

\title{Reproducible Evaluation as the Substrate for Safe Prompt Compression in Production Agents}

\author{%
  Prabal Vashisht \\
  Mx ML\\
  DoorDash Inc.\\
  \texttt{prabal.vashisht@doordash.com} \\
  \And
  Zheng Hu \\
  Mx ML\\
  DoorDash Inc.\\
  \texttt{zheng.hu@doordash.com} \\
  \And
  Ying Yang \\
  Mx ML\\
  DoorDash Inc.\\
  \texttt{ying.yang@doordash.com} \\
}

\begin{document}

\maketitle

\begin{abstract}
Evaluating conversational agents that act on live, mutable enterprise data is hard: replayed sessions drift as backend state changes, and a benchmark built today can silently go stale as the data it describes moves on. We present an offline evaluation harness for a production conversational agent that stays reproducible despite the live data it operates on changing, and that can be regenerated on demand once it goes stale. We then use this harness as the fitness function for GEPA-based prompt optimization under a correctness gate that guards against reward hacking. Applied to a production merchant-facing agent, correctness-gated compression shrinks the agent's full prompt surface by
3.5--7.6\% without regressing held-out task correctness, consistently
across three independent subject-model families --- and for one, correctness measurably improves.
\end{abstract}

\section{Introduction}

Conversational agents increasingly serve as the interface between users and
the software systems that support their work --- answering questions grounded
in live data \citep{lewis2020rag}, and, following the shift toward
tool-augmented and agentic LLMs
\citep{yao2022react,schick2023toolformer,wang2024survey}, increasingly taking
actions on that data rather than only reporting on it.

In practice, teams do not build such an agent by handing an optimizer a blank
prompt and a training set. They first fix the product spec and tool
contracts --- required wording, forbidden wording, which tool answers which
question \citep{schluntz2024building} --- so the agent's behavior stays
legible and reviewable. Only once this spec-complete agent exists does the
natural question become: is it reliable, and is it efficient? This paper is a
blueprint for that starting point: given a production agent whose spec and
tools are fixed, how do we make its evaluation trustworthy and its prompt
leaner, without re-deriving its behavior from scratch?

Reliability is harder to establish here than for static NLP tasks: the
ground truth an agent is queried against is not fixed --- a sales question has
a different correct answer daily, and a write action changes the state the
next eval run would read. Naive session replay against a live backend
therefore produces incomparable results, and scoring a write tool's text
misses the point --- what matters is the state it leaves behind. A benchmark
can also go stale after the fact, as the entities it describes move on (a menu
reprices, a payout settles), drifting from the system it describes
independent of any model change. We treat this as routine: because generation
and evaluation share one schema, a stale benchmark can be regenerated on
demand from current backend state.

Prior work addresses pieces of this. Static benchmarks like ToolLLM
\citep{qin2023toolllm} assume a frozen corpus; interactive-environment
benchmarks like AgentBench \citep{liu2023agentbench}, WebArena
\citep{zhou2023webarena}, and $\tau$-bench \citep{yao2024tau} score end
state, but inside simulated environments rather than live, drifting
production data. Dynabench \citep{kiela2021dynabench} and LiveBench
\citep{white2024livebench} refresh benchmarks against saturation and
contamination, respectively; our staleness instead comes from the referent
data moving, which calls for regeneration rather than continual sourcing.
LLM-as-judge scoring \citep{zheng2023judging} handles subjective wording but
not reproducibility itself, and prompt optimizers --- APE \citep{zhou2022ape},
DSPy \citep{khattab2023dspy}, GEPA \citep{agrawal2025gepa} --- treat their
evaluation signal as given rather than addressing what makes it reproducible
or fresh.

Our harness matches a reproducibility mechanism to each tool's data shape:
staging the clock for reads immutable once settled, freezing the observed
payload for current-state-dependent reads, and seed-then-read-back for tools
that write. Archetype-stratified generation covers each skill's capability
surface, and deterministic plus LLM-rubric judges \citep{zheng2023judging}
score correctness and policy compliance.

Because this harness yields a stable scalar signal, it can also serve as
GEPA's \citep{agrawal2025gepa} fitness function. Published GEPA use cases grow
a minimal seed prompt via reflection; production agents start spec-complete,
so optimization must instead \emph{compress} without regressing. Shorter
prompts directly reduce inference latency and serving cost
\citep{jiang2023llmlingua}, so token count is worth optimizing even once
correctness is already met. We use a lexicographic token-count objective
that only breaks exact performance ties, never trading correctness for size,
guarded by a hard gate against any rule-violating candidate --- a
prompt-optimization analogue of reward hacking in reinforcement learning
\citep{skalse2022reward}. We validate this on a merchant-facing conversational
agent already in production: across three independent subject-model
families, compression under our gate reduces the agent's full prompt surface
by 3.5--7.6\% without regressing held-out task correctness on a confirmatory
held-out set --- and for one family, correctness measurably improves ---
evidence that prompt size and task correctness need not trade off against
each other.

Our contributions: (1) a sandbox that generates and evaluates benchmark data
against live production state, matching each tool's data-mutability to a
reproducibility mechanism and scoring write-tools on end state rather than
transcript text; (2) a generation strategy that can be re-run at any point to
refresh a benchmark once its underlying production data has moved on; and (3)
a token-efficiency method that uses this harness as a correctness-gated
fitness function for prompt compression.

\section{Methodology}
\subsection{Preliminaries}

Following GEPA \citep{agrawal2025gepa} and the compound AI system framing it
builds on \citep{zaharia2024compound}, we treat the agent as a compound
system $\Phi = (M, C, X, Y)$: modules $M = \langle M_1, \dots,
M_{|M|}\rangle$, each $M_i = (\pi_i, \theta_i, X_i, Y_i)$ with its own prompt
and weights, orchestrated by control flow $C$.

A skills-based agent's $M$ centers on a core module $M_\text{core}$, whose
prompt $\pi_\text{core}$ is composed from a skill registry $S = \{s_1,
\dots, s_n\}$ --- following the Agent Skills convention
\citep{anthropic2025skills} of packaging instructions and resources under a
discoverable name and description --- $s_j = (K_j, T_j)$ pairing
instruction text $K_j$ with tools $T_j$. $C$ maintains an active subset $A
\subseteq S$ per turn, so the model sees
\[
\pi_\text{core} = \pi_\text{base} \oplus \bigcup_{s_j \in A} K_j, \qquad
T_A = \bigcup_{s_j \in A} T_j,
\]
where $\pi_\text{base}$ is the agent's system prompt, independent of any
skill. It is what our contributions act on: the sandbox is skill-agnostic
infrastructure for evaluating safely against production data, and
compression jointly rephrases every prompt ($\pi_\text{base}$, $K_j$)
together with the tools' description text.

\subsection{Sandbox: Reproducible Production State}
\label{sec:sandbox}

A benchmark datapoint is a triple $d = (x, y^\star, e)$: a task $x$
exercising skills $A \subseteq S$, its ground truth $y^\star$, and the
production-state slice $e \in \mathcal{E}$ needed to reproduce it. Each tool
has a mutability class $\mu \in \{\textit{settled}, \textit{current},
\textit{write}\}$, mapped to a reproducibility mechanism $\rho(\mu)$: stage
the clock for settled reads, freeze the payload for current-state reads, or
seed-then-read-back for writes --- letting $e$ replay identically despite
$\mathcal{E}$ drifting.

$\rho$ only covers $d$'s staged scope $A$; a call routed outside it hits
live, unstaged state, and --- unlike a failed seed, hard-gated to 0 ---
goes unflagged, reading as an ordinary wrong answer.

Evaluation runs $M_\text{core}$ against $d$ under $\rho(\mu)$:
$\text{Eval}(y, y^\star) \in [0,1]$ for reads, or $\text{Eval}(e_\text{post},
e^\star_\text{post})$ over sandboxed end state for writes, since a write
tool is graded on what it leaves behind, not its text. Score is
$\text{Score}(P; D') = \frac{1}{|D'|}\sum_{d \in D'} \text{Eval}(\cdot)$ for a
prompt component $P$ substituted into $M_\text{core}$ and evaluated over
datapoints $D'$.

\subsection{Generation: Grounding Datapoints in Achievable Tool-Call Traces}
\label{sec:generation}

A task $x$ only tests skills $A \subseteq S$'s product spec if it's
answerable with $T_A = \bigcup_{s_j \in A} T_j$. Two properties gate
admission. First, \emph{existence}: a generator agent, scoped to
$\pi_\text{core}$ and $T_A$ --- exactly what $M_\text{core}$ sees on that
turn --- attempts $(x, \tau, y^\star)$ against a sampled $e$, captured by
applying the same $\rho(\mu)$ from \S\ref{sec:sandbox} so the state $\tau$
observes replays identically at evaluation time, where $\tau = (t_1, \dots,
t_r)$, $t_i \in T_A$, is the trace deriving $y^\star$. A slot is kept only
if $\tau$ executes end to end and $y^\star$ follows from it; otherwise it's
discarded and resampled against a fresh $e$.

Second, \emph{sense}: since $(x, \tau, y^\star)$ was produced by an agent,
not hand-authored, a real trace doesn't guarantee a good test --- $x$ can
still be ill-posed, or $y^\star$ technically derived but not what the spec
calls for. A human reviews every surviving datapoint before admission to
$D$, confirming $x$ is well-posed, $\tau$ reflects the intended capability
rather than an incidental path, and $y^\star$ matches the spec, not just
the generator's phrasing. Archetypes stratify generation across each
skill's capability surface so $D$ doesn't collapse onto a few easy shapes.

Because generation is this repeatable pipeline over $e$, not a one-off
authoring pass, it can be re-run at any later point $u$ against $e_u$ ---
trace, human review, and all --- to produce a fresh $D'$ once the original
goes stale.

\subsection{Correctness-Gated Prompt Compression}
\label{sec:compression}

GEPA optimizes one joint candidate $c$ spanning the registry's textual
surface: the base system prompt $\pi_\text{base}$, every skill's $K_j$,
plus the description text of every tool in $\bigcup_j T_j$. Compressing
any one of these alone can only relocate text duplicated elsewhere, not
remove it --- coordinated rephrasing across the whole surface is what
actually shrinks it. Let $D = \bigcup_j D_j$ pool datapoints across skills,
partitioned into disjoint $D^\text{train}$, $D^\text{val}$, and
$D^\text{test}$ (\S\ref{sec:results}), and $|c|_\text{tok}$ be $c$'s total
token count; fitness is
\[
f(c) =
\begin{cases}
  0 & c \text{ violates a mandatory rule (critical gate)} \\
  \text{Score}(c; D^\text{val}) - \epsilon \cdot |c|_\text{tok} / |D^\text{val}| & \text{otherwise.}
\end{cases}
\]
The gate is a hard zero, not a soft penalty, because compression would
otherwise reward deleting a mandatory instruction whenever few datapoints
exercise it --- a prompt-optimization analogue of reward hacking
\citep{skalse2022reward}. $\epsilon$ is sized to only break exact
$\text{Score}$ ties among rule-compliant candidates in the final selection,
never trading correctness for size.

GEPA edits every component of $c$ each iteration --- not one at a time,
which can't coordinate such a move --- from $D^\text{train}$ failures.
Mutation parents are sampled from a Pareto frontier that unions candidates
non-dominated on a per-datapoint basis with candidates non-dominated on
aggregate per-objective scores --- giving the search room to explore
genuine correctness/size trade-offs before committing. The candidate
ultimately reported, though, is whichever maximizes aggregate $f$ over
$D^\text{val}$ alone, so the trade-offs the frontier explores never
determine the final answer; a held-out $D^\text{test}$, disjoint from
both, reports the honest result. Unlike typical GEPA, which grows $c$
from a minimal seed, $c$ starts spec-complete, so every accepted mutation
compresses rather than adds.

\section{Results}
\label{sec:results}

We apply this harness to a merchant-facing conversational agent in
production, optimizing every textual prompt component of the agent
--- the base system prompt $\pi_\text{base}$, every skill's $K_j$,
and the description text of every tool --- jointly as one candidate
$c$ (\S\ref{sec:compression}) against the full generated dataset.

\textbf{Dataset.} Following \S\ref{sec:sandbox}--\ref{sec:generation}, we
generate a train/val split stratified by skill--archetype pair across
10 skills at seed 42: 302 datapoints for training and 99 for validation.
A 196-point held-out set $D^\text{test}$, disjoint from train and
val and also stratified by skill--archetype pair, reports the honest
result (\S\ref{sec:compression}).

\textbf{Compression run.} We run GEPA independently for three
subject-model families --- Anthropic (\texttt{claude-sonnet-4.6}), Gemini
(\texttt{gemini-3.7-flash}), and OpenAI (\texttt{gpt-5.6-sol}) --- each
optimizing the same joint candidate $c$ (system prompt $\pi_\text{base}$,
all ten skill bodies $K_j$, and shared tool descriptions; 36{,}336-token
seed) at a rollout budget of 800. Fitness follows the $f(c)$ defined in \S\ref{sec:compression}
($\epsilon = 10^{-6}$), with the critical-rule gate checked against
every skill's mandatory rules; candidates are additionally rejected by
a no-harm gate that zeroes any candidate whose per-skill mean falls
more than 0.02 below the seed's. The Pareto frontier is
\texttt{hybrid} --- maintaining non-dominated candidates from two
complementary views, per-datapoint performance and aggregate objective
scores, rather than either alone --- and the module selector is set to
\texttt{all}, editing every component within the same reflection step.
Seed and winner are each scored on $D^\text{test}$ as the mean of 5
rollouts per datapoint, controlling for sampling noise;
Table~\ref{tab:compression} reports the paired difference with a
10,000-sample bootstrap 95\% confidence interval.

\begin{table}[t]
\centering
\small
\begin{tabular}{lrrc}
\toprule
Family & Tokens (seed$\to$winner) & Compression & $D^\text{test}$ $\Delta$ (95\% CI) \\
\midrule
Anthropic & 36{,}336 $\to$ 33{,}567 & $-7.6\%$ & $+0.118$ [$+0.073$, $+0.164$] \\
Gemini    & 36{,}336 $\to$ 34{,}380 & $-5.4\%$ & $+0.016$ [$-0.001$, $+0.035$] \\
OpenAI    & 36{,}336 $\to$ 35{,}077 & $-3.5\%$ & $+0.006$ [$-0.007$, $+0.021$] \\
\bottomrule
\end{tabular}
\caption{GEPA compression results per subject-model family. $\Delta$ is
the paired difference (winner $-$ seed) on $D^\text{test}$ ($n=196$),
mean of 5 rollouts per datapoint, with a 10,000-sample bootstrap 95\%
confidence interval.}
\label{tab:compression}
\end{table}

Only Anthropic's confidence interval excludes zero: its winner is both
7.6\% smaller and a real, statistically distinguishable improvement over
its seed. Gemini's and OpenAI's intervals include zero --- their winners
are smaller with no detectable change in quality, not a proven gain. Read
together, compression does not cost correctness for any family, and for
one it recovers real headroom the seed's extra tokens did not provide.

\section{Conclusion}

We presented an offline evaluation harness that grounds benchmark
datapoints in reproducible production state and real tool-call traces,
paired with a correctness-gated compression method that treats a compound
agent's entire textual surface as one candidate optimized under a hard
gate against reward hacking. Applied to a merchant-facing agent already in production, this let GEPA
shrink the agent's prompt by 3.5--7.6\% without regressing held-out
correctness on a confirmatory held-out set, across three independent
subject-model families (\S\ref{sec:results}) --- and for one family,
correctness measurably improves --- evidence that prompt size and task
correctness need not trade off against each other.

The gain's size varies by subject model, so its benefit appears to depend
on the model optimized, not solely on the harness. And since our
benchmark is a frozen snapshot, we have not yet measured how a compressed
candidate's advantage decays as production drifts, despite the harness
being designed to regenerate on demand (\S\ref{sec:generation}).

\newpage

\bibliographystyle{plainnat}
\bibliography{references}

@inproceedings{lewis2020rag,
  title={Retrieval-Augmented Generation for Knowledge-Intensive NLP Tasks},
  author={Lewis, Patrick and Perez, Ethan and Piktus, Aleksandra and Petroni, Fabio and Karpukhin, Vladimir and Goyal, Naman and K{\"u}ttler, Heinrich and Lewis, Mike and Yih, Wen-tau and Rockt{\"a}schel, Tim and Riedel, Sebastian and Kiela, Douwe},
  booktitle={Advances in Neural Information Processing Systems},
  volume={33},
  pages={9459--9474},
  year={2020}
}

@inproceedings{yao2022react,
  title={ReAct: Synergizing Reasoning and Acting in Language Models},
  author={Yao, Shunyu and Zhao, Jeffrey and Yu, Dian and Du, Nan and Shafran, Izhak and Narasimhan, Karthik and Cao, Yuan},
  booktitle={International Conference on Learning Representations},
  year={2023},
  note={arXiv:2210.03629}
}

@inproceedings{schick2023toolformer,
  title={Toolformer: Language Models Can Teach Themselves to Use Tools},
  author={Schick, Timo and Dwivedi-Yu, Jane and Dess{\`i}, Roberto and Raileanu, Roberta and Lomeli, Maria and Zettlemoyer, Luke and Cancedda, Nicola and Scialom, Thomas},
  booktitle={Advances in Neural Information Processing Systems},
  year={2023},
  note={arXiv:2302.04761}
}

@article{wang2024survey,
  title={A Survey on Large Language Model based Autonomous Agents},
  author={Wang, Lei and Ma, Chen and Feng, Xueyang and Zhang, Zeyu and Yang, Hao and Zhang, Jingsen and Chen, Zhiyuan and Tang, Jiakai and Chen, Xu and Lin, Yankai and others},
  journal={Frontiers of Computer Science},
  volume={18},
  number={6},
  year={2024},
  note={arXiv:2308.11432}
}

@misc{schluntz2024building,
  title={Building Effective Agents},
  author={Schluntz, Erik and Zhang, Barry},
  year={2024},
  howpublished={Anthropic Engineering Blog, Dec.\ 19, 2024},
  url={https://www.anthropic.com/engineering/building-effective-agents}
}

@inproceedings{liu2023agentbench,
  title={AgentBench: Evaluating LLMs as Agents},
  author={Liu, Xiao and Yu, Hao and Zhang, Hanchen and Xu, Yifan and Lei, Xuanyu and Lai, Hanyu and Gu, Yu and Ding, Hangliang and Men, Kaiwen and Yang, Kejuan and Zhang, Shudan and Deng, Xiang and Zeng, Aohan and Du, Zhengxiao and Zhang, Chenhui and Shen, Sheng and Zhang, Tianjun and Su, Yu and Sun, Huan and Huang, Minlie and Dong, Yuxiao and Tang, Jie},
  booktitle={International Conference on Learning Representations},
  year={2024},
  note={arXiv:2308.03688}
}

@inproceedings{qin2023toolllm,
  title={ToolLLM: Facilitating Large Language Models to Master 16000+ Real-world APIs},
  author={Qin, Yujia and Liang, Shihao and Ye, Yining and Zhu, Kunlun and Yan, Lan and Lu, Yaxi and Lin, Yankai and Cong, Xin and Tang, Xiangru and Qian, Bill and Zhao, Sihan and Hong, Lauren and Tian, Runchu and Xie, Ruobing and Zhou, Jie and Gerstein, Mark and Li, Dahai and Liu, Zhiyuan and Sun, Maosong},
  booktitle={International Conference on Learning Representations},
  year={2024},
  note={Spotlight; arXiv:2307.16789}
}

@article{yao2024tau,
  title={$\tau$-bench: A Benchmark for Tool-Agent-User Interaction in Real-World Domains},
  author={Yao, Shunyu and Shinn, Noah and Razavi, Pedram and Narasimhan, Karthik},
  journal={arXiv preprint arXiv:2406.12045},
  year={2024}
}

@inproceedings{zheng2023judging,
  title={Judging LLM-as-a-Judge with MT-Bench and Chatbot Arena},
  author={Zheng, Lianmin and Chiang, Wei-Lin and Sheng, Ying and Zhuang, Siyuan and Wu, Zhanghao and others},
  booktitle={Advances in Neural Information Processing Systems},
  year={2023},
  note={Datasets and Benchmarks Track; arXiv:2306.05685}
}

@inproceedings{zhou2022ape,
  title={Large Language Models Are Human-Level Prompt Engineers},
  author={Zhou, Yongchao and Muresanu, Andrei Ioan and Han, Ziwen and Paster, Keiran and Pitis, Silviu and Chan, Harris and Ba, Jimmy},
  booktitle={International Conference on Learning Representations},
  year={2023},
  note={arXiv:2211.01910}
}

@inproceedings{khattab2023dspy,
  title={DSPy: Compiling Declarative Language Model Calls into State-of-the-Art Pipelines},
  author={Khattab, Omar and Singhvi, Arnav and Maheshwari, Paridhi and Zhang, Zhiyuan and Santhanam, Keshav and Vardhamanan, Sri and Haq, Saiful and Sharma, Ashutosh and Joshi, Thomas T. and Moazam, Hanna and Miller, Heather and Zaharia, Matei and Potts, Christopher},
  booktitle={International Conference on Learning Representations},
  year={2024}
}

@inproceedings{skalse2022reward,
  title={Defining and Characterizing Reward Hacking},
  author={Skalse, Joar and Howe, Nikolaus and Krasheninnikov, Dmitrii and Krueger, David},
  booktitle={Advances in Neural Information Processing Systems},
  year={2022},
  note={arXiv:2209.13085}
}

@article{agrawal2025gepa,
  title={GEPA: Reflective Prompt Evolution Can Outperform Reinforcement Learning},
  author={Agrawal, Lakshya A and Tan, Shangyin and Soylu, Dilara and Ziems, Noah and Khare, Rishi and Opsahl-Ong, Krista and Singhvi, Arnav and Shandilya, Herumb and Ryan, Michael J and Jiang, Meng and Potts, Christopher and Sen, Koushik and Dimakis, Alexandros G. and Stoica, Ion and Klein, Dan and Zaharia, Matei and Khattab, Omar},
  journal={arXiv preprint arXiv:2507.19457},
  year={2025},
  note={ICLR 2026 Oral}
}

@inproceedings{zhou2023webarena,
  title={WebArena: A Realistic Web Environment for Building Autonomous Agents},
  author={Zhou, Shuyan and Xu, Frank F. and Zhu, Hao and Zhou, Xuhui and Lo, Robert and Sridhar, Abishek and Cheng, Xianyi and Ou, Tianyue and Bisk, Yonatan and Fried, Daniel and Alon, Uri and Neubig, Graham},
  booktitle={International Conference on Learning Representations},
  year={2024},
  note={arXiv:2307.13854}
}

@inproceedings{white2024livebench,
  title={LiveBench: A Challenging, Contamination-Limited LLM Benchmark},
  author={White, Colin and Dooley, Samuel and Roberts, Manley and Pal, Arka and Feuer, Ben and Jain, Siddhartha and Shwartz-Ziv, Ravid and Jain, Neel and Saifullah, Khalid and Dey, Sreemanti and Agrawal, Shubh and Sandha, Sandeep Singh and Naidu, Siddartha and Hegde, Chinmay and LeCun, Yann and Goldstein, Tom and Neiswanger, Willie and Goldblum, Micah},
  booktitle={International Conference on Learning Representations},
  year={2025},
  note={Spotlight; arXiv:2406.19314}
}

@inproceedings{kiela2021dynabench,
  title={Dynabench: Rethinking Benchmarking in NLP},
  author={Kiela, Douwe and Bartolo, Max and Nie, Yixin and Kaushik, Divyansh and Geiger, Atticus and Wu, Zhengxuan and Vidgen, Bertie and Prasad, Grusha and Singh, Amanpreet and Ringshia, Pratik and Ma, Zhiyi and Thrush, Tristan and Riedel, Sebastian and Waseem, Zeerak and Stenetorp, Pontus and Jia, Robin and Bansal, Mohit and Potts, Christopher and Williams, Adina},
  booktitle={Proceedings of NAACL-HLT},
  year={2021}
}

@misc{zaharia2024compound,
  title={The Shift from Models to Compound AI Systems},
  author={Zaharia, Matei and Khattab, Omar and Chen, Lingjiao and Davis, Jared Quincy and Miller, Heather and Potts, Christopher and Zou, James and Carbin, Michael and Frankle, Jonathan and Rao, Naveen and Ghodsi, Ali},
  year={2024},
  howpublished={Berkeley Artificial Intelligence Research (BAIR) Blog, Feb.\ 18, 2024},
  url={https://bair.berkeley.edu/blog/2024/02/18/compound-ai-systems/}
}

@misc{anthropic2025skills,
  title={Equipping Agents for the Real World with Agent Skills},
  author={{Anthropic}},
  year={2025},
  howpublished={Anthropic Engineering Blog},
  url={https://www.anthropic.com/engineering/equipping-agents-for-the-real-world-with-agent-skills}
}

@inproceedings{jiang2023llmlingua,
  title     = {{LLMLingua}: Compressing Prompts for Accelerated Inference of Large Language Models},
  author    = {Jiang, Huiqiang and Wu, Qianhui and Lin, Chin-Yew and Yang, Yuqing and Qiu, Lili},
  booktitle = {Proceedings of the 2023 Conference on Empirical Methods in Natural Language Processing},
  year      = {2023},
  eprint    = {2310.05736},
}

\clearpage




\appendix

\section{Example Datapoint}
\label{app:datapoint}

To make the harness's data format concrete, the listing below reproduces
one datapoint from the \texttt{sales} skill's generated dataset
(\S\ref{sec:sandbox}--\ref{sec:generation}), with the store identifier and
the generation tool call's arguments/output redacted.

\begin{footnotesize}
\begin{verbatim}
datapoint_id: sales-fcb32a3d
tags: [sales]
merchant:
  store_id: <STORE_ID>       # synthetic eval-mode test merchant
  eval_mode: true
task:
  description: >
    Can you show me my daily revenue trend from April 13 to April 19,
    2026 (inclusive of both dates) as a line chart?
  expected_tool_calls: [query_sales_data, render_chart]
environment:
  as_of_datetime: 2026-04-28T10:59:37-07:00
simulator:
  kind: single_shot
  opening_message: >
    Can you show me my daily revenue trend from April 13 to April 19,
    2026 (inclusive of both dates) as a line chart?
evaluators:
  - type: dollar_recall
    expected_final_response: >
      Your total revenue from April 13 to April 19, 2026 was $352.28
      across 18 orders, with an average order value of $19.57. ...
    must_have_figures: ["352.28"]
score_aggregation: min
generation_tool_calls:
  - name: query_sales_data
    args: <REDACTED>
    output: <REDACTED>
\end{verbatim}
\end{footnotesize}

The task $x$ is the merchant's natural-language question together with
the \texttt{merchant} context; the expected outcome $y^*$ is the
\texttt{dollar\_recall} evaluator's expected response and required
figures; the existence witness $\tau$ is the
\texttt{generation\_tool\_calls} trace --- here a single real
\texttt{query\_sales\_data} call against production data, captured at
\texttt{as\_of\_datetime}. Note that \texttt{render\_chart} appears in
\texttt{expected\_tool\_calls} but not in the captured trace: only calls
that fetch state backing the graded answer are retained as the witness,
not every call in the full interaction.

\section{Reward-Hacking Near-Miss}
\label{app:reward-hacking}

During search, one intermediate candidate's only edit was dropping a
frozen \texttt{store\_id\_handling} block from \texttt{sales\_skill\_body}:

\begin{footnotesize}
\begin{verbatim}
-## Store ID
-
-- StoreLevel: pass `store_id` from Current Entity.
-- BusinessLevel: **omit** `store_id` for business-wide results. Never
-  use `business_id` as `store_id`.
-
 ## request_json Parameters
\end{verbatim}
\end{footnotesize}

This candidate's own no-harm check correctly zeroed it. But a later
candidate in the same lineage inherited the deletion and was not caught
the same way --- the no-harm gate is computed per-iteration against
whichever batch is sampled that round, so a rule a given batch never
exercises can slip through. Independently, the text-coverage judge checks
for the rule's text anywhere in the candidate, not per-batch --- and here
it found the same instruction duplicated earlier in the skill body and
passed the candidate without checking that the specific copy it needed
was the one just deleted. Only a broader diagnostic pass, checking the \texttt{sales} skill
specifically, surfaced the drop (0.90 $\to$ 0.00) --- after the fact,
not as part of the standard selection pipeline. This near-miss is
exactly the gap the hard critical-rule gate
(\S\ref{sec:compression}) closes going forward: unlike the no-harm
gate's batch-dependent sampling or the text-coverage judge's static,
duplicate-blind text check, the critical-rule gate runs
deterministically against every candidate's actual rule compliance,
independent of which batch happened to be sampled or which text
happens to appear elsewhere in the skill body.

\section{Sample Compression}
\label{app:compression}

Below is a short excerpt from the \texttt{financials} skill body,
before and after the Anthropic family's winning edit,

\textbf{Seed:}
\begin{footnotesize}
\begin{verbatim}
## Overview
Help merchants with two related financial topics:
- **Refunds & adjustments** -- orders where money was returned to the
  merchant (dispute refunds or adjustment credits)
- **Payouts & fees** -- deposit history, payout schedules, commission
  rates, and fee structure across stores

## Display Rules (MANDATORY)
**NEVER output text lists.** Always use frontend tools to display data.
- `<TOOL_REFUNDS>` returns pre-formatted table data -> pass directly to
  the table renderer
- Never output markdown tables or plain text lists
\end{verbatim}
\end{footnotesize}

\textbf{Winner:}
\begin{footnotesize}
\begin{verbatim}
## Overview
Help merchants with: **Refunds & adjustments** (money returned to
merchant) and **Payouts & fees** (deposit history, schedules,
commission rates, fee structure).

## Display Rules (MANDATORY)
**NEVER output text lists or markdown tables.** Always use frontend
tools.
- `<TOOL_REFUNDS>` -> pass table data directly to the table renderer
\end{verbatim}
\end{footnotesize}

The two-bullet Overview collapses into one sentence, and the Display
Rules section drops a redundant instruction --- \emph{``NEVER output
text lists''} and \emph{``Never output markdown tables or plain text
lists''} merge into a single rule --- without dropping any distinct
constraint --- exactly the kind of lossless rewording the harness is
designed to reward.

\newpage

\end{document}